# Semantic Segmentation using Adversarial Networks

**Pauline Luc**
Facebook AI Research
Paris, France
paulineluc@fb.com

**Camille Couprie**
Facebook AI Research
Paris, France
coupriec@fb.com

**Soumith Chintala**
Facebook AI Research
New York, USA
soumith@fb.com

**Jakob Verbeek**
INRIA, Laboratoire Jean Kuntzmann
Grenoble, France
firstname.lastname@inria.fr

## Abstract

Adversarial training has been shown to produce state of the art results for generative image modeling. In this paper we propose an adversarial training approach to train semantic segmentation models. We train a convolutional semantic segmentation network along with an adversarial network that discriminates segmentation maps coming either from the ground truth or from the segmentation network. The motivation for our approach is that it can detect and correct higher-order inconsistencies between ground truth segmentation maps and the ones produced by the segmentation net. Our experiments show that our adversarial training approach leads to improved accuracy on the Stanford Background and PASCAL VOC 2012 datasets.

## 1 Introduction

Semantic segmentation is a visual scene understanding task formulated as a dense labeling problem, where the goal is to predict a category label at each pixel in the input image. Current state-of-the-art methods [2, 15, 16, 21] rely on convolutional neural network (CNN) approaches, following early work using CNNs for this task by Grangier *et al*. in 2009 [11] and Farabet *et al*. [7] in 2013. Despite many differences in the CNN architectures, a common property across all these approaches is that all label variables are predicted independently from each other. This is the case at least during training; various post-processing approaches have been explored to reinforce spatial contiguity in the output label maps since the independent prediction model does not capture this explicitly.

Conditional Markov random fields (CRFs) are one of the most effective approaches to enforce spatial contiguity in the output label maps. The CNN-based approaches mentioned above can be used to define unary potentials. For certain classes of pairwise potentials, mean-field inference in fully-connected CRFs with millions of variables is tractable using recent filter-based techniques[14]. Such fully-connected CRFs have been found extremely effective in practice to recover fine details in the output maps. Moreover, using a recurrent neural network formulation [30, 35] of the mean-field iterations, it is possible to train the CNN underlying the unary potentials in an integrated manner that takes into account the CRF inference during training. It has also been shown that a rich class of pairwise potentials can be learned using CNN techniques in locally connected CRFs [15].

Despite these advances, the work discussed above is limited to the use of pairwise CRF models. Higher-order potentials, however, have also been observed to be effective, for example robust higher-order terms based on label consistency across superpixels [13]. Recent work [1] has shown how specific classes of higher order potentials can be integrated in CNN-based segmentation models. While the parameters of these higher-order potentials can be learned, they are limited in number.

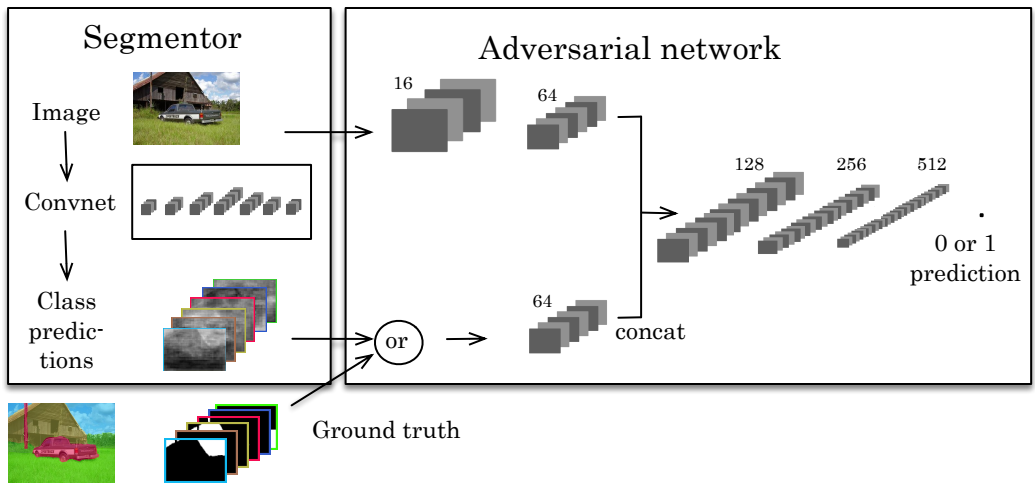

Figure 1: Overview of the proposed approach. Left: segmentation net takes RGB image as input, and produces per-pixel class predictions. Right: Adversarial net takes label map as input and produces class label (1=ground truth, or 0=synthetic). Adversarial optionally also takes RGB image as input.

We are interested in enforcing higher-order consistency without being limited to a very specific class of high-order potentials. Instead of seeking to directly integrate higher-order potentials in a CRF model, we explore an approach based on adversarial training inspired by the generative adversarial network (GAN) approach of Goodfellow *et al*. [9]. To this end we optimize an objective function that combines a conventional multi-class cross-entropy loss with an adversarial term. The adversarial term encourages the segmentation model to produce label maps that cannot be distinguished from ground-truth ones by an adversarial binary classification model. Since the adversarial model can assess the joint configuration of many label variables, it can enforce forms of higher-order consistency that cannot be enforced using pair-wise terms, nor measured by a per-pixel cross-entropy loss.

The contributions of our work are the following:

1. We present, to the best of our knowledge, the first application of adversarial training to semantic segmentation.
2. The adversarial training approach enforces long-range spatial label contiguity, without adding complexity to the model used at test time.
3. Our experimental results on the Stanford Background and PASCAL VOC 2012 dataset show that our approach leads to improved labeling accuracy.

We discuss related work on adversarial training approaches, as well as recent CNN-based segmentation models, in Section 2. After presenting our adversarial training approach and network architectures in Section 3, we present experimental results in Section 4. We conclude with a discussion in Section 5.

## 2 Related work

**Adversarial learning.** Goodfellow *et al*. [9] proposed an adversarial approach to learn deep generative models. Their generative adversarial networks (GANs) take samples $z$ from a fixed (*e.g.* standard Gaussian) distribution $p_z(z)$, and transform them by a deterministic differentiable deep network $g(\cdot)$ to approximate the distribution of training samples $x$. Note that the distribution $p_x(\cdot)$ over $x$ induced by $g(\cdot)$ and $p_z(\cdot)$ is intractable to evaluate due to the integration over $z$ across the highly non-linear deep net. An adversarial network is used to define a loss function which sidesteps the need to explicitly evaluate or approximate $p_x(\cdot)$. The adversarial model is trained to optimally discriminate samples from the empirical data distribution and samples from the deep generative model.

The generative model is concurrently trained to minimize accuracy of the adversarial, which provably drives the generative model to approximate the distribution of the training data. The adversarial net can be interpreted as a "variational" loss function, in the sense that the loss function of the generative model is defined by auxiliary parameters that are not part of the generative model.

In follow-up work, Radford *et al*. [25] present a number of architectural design choices that enable stable training of generative models that are able to synthesize realistic images. Similar to Goodfellow *et al*. [9], they use deep "deconvolutional" networks $g(\cdot)$ that progressively construct the image by up-sampling, using essentially a reverse CNN architecture. Denton *et al*. [4] use a Laplacian pyramid approach to learn a sequence of GAN models that successively generate images with finer details. Extensions of GANs for conditional modeling have been explored, *e.g.* for image tag prediction [19], face image generation conditioned on attributes [8], and for caption-based image synthesis [26].

Deep conditional generative models have also been defined in a non-stochastic manner, where for a given conditioning variable a single deterministic output is generated. For example, Dosovitskiy *et al*. [5] developed deep generative image models where the conditioning variables encode the object class, viewpoint, and color-related transformations. In this case a conventional regression loss can be used, since inference or integration on the conditioning variables is not needed. Dosovitskiy *et al*. train their models using an $\ell_2$ regression loss on the target images.

In other examples the conditioning variable takes the form of one or more input images. Mathieu *et al*. [18] considered the problem of predicting the next frame in video given several preceding frames. Pathak *et al*. [22] considered the problem of image inpainting, where the missing part of the images has to be predicted from the observed part. Such models are closely related to deep convolutional semantic segmentation models that deterministically produce a label probability map, conditioned on an input RGB image. In the latter two cases, a regression loss in combined with an adversarial loss term. The motivation in both cases is that per-pixel regression losses typically result in too blurry outputs, since they do not for higher-order regularities in the output. Since the adversarial net has access to large portions or the entire output image, it can be interpreted as a learned higher-order loss, which obviates the need to manually design higher-order loss terms. The work of Tarlow and Zemel [33] is related to this approach as they also suggested to learn with higher-order loss terms, while not including such higher-order terms in the predictive model to ensure efficient prediction. Several authors have shown that images on which convolutional classification networks produce confident but incorrect predictions can be found by manipulating natural images in a human-imperceptible manner [32], or by synthesizing non-natural images [20]. This is related to adversarial training in the sense that they seek to reduce the CNN performance by perturbing the input, in GANs these perturbations are further back-propagated through the generative network to improve its performance.

**Semantic segmentation.** While early CNN-based semantic segmentation approaches were explicitly passing image patches through the CNN, see *e.g.* [7], current state-of-the-art method indifferently use a fully convolutional approach [16]. This is far more efficient, since it avoids redundant computation of low-level filters many times on pixels in overlapping patches. Typical architectures involve a number of pooling steps, which can increase the receptive field size rapidly after several steps. As a result, however, the resolution of the output maps reduces, which means that a low-resolution label map is obtained. To address this issue, the signal can be up-sampled using bi-linear interpolation, or learned up-sampling filters [16, 21, 27]. Alternatively, one can use dilated convolutions to increase the receptive field size without losing resolution [2, 34], skip connections to earlier high-resolution layers [16, 27], or multi-resolution networks [29, 36].

Most work that combines CNN unary label predictions with CRFs is based on models with pairwise or higher-order terms with few trainable parameters [1, 30, 35]. An exception is the work of Lin *et al*. [15] which uses a second CNN to learn data dependent pairwise terms. Another approach that exploits high-capacity trainable models to drive long-range label interactions is to use recurrent networks [23], where each iteration maps the input image and current label map to a new label map.

In comparison to these previous approaches our work has the following merits: (i) The adversarial model has a high capacity, and is thus flexible enough to detect mismatches in a wide range of higher-order statistics between the model predictions and the ground-truth, without having to manually define these. (ii) Once trained, our model is efficient since it does not involve any higher-order terms or recurrence in the model itself.

## 3 Adversarial training for semantic segmentation networks

We describe our general framework for adversarial training of semantic segmentation models in Section 3.1. We present the architectures used in our experiments in Section 3.2.

### 3.1 Adversarial training

We propose to use a hybrid loss function that is a weighted sum of two terms. The first is a multi-class cross-entropy term that encourages the segmentation model to predict the right class label at each pixel location independently. This loss is standard in state-of-the-art semantic segmentation models, see *e.g.* [2, 15, 16, 21]. We use $s(\boldsymbol{x})$ to denote the class probability map over $C$ classes of size $H \times W \times C$ that the segmentation model produces given an input RGB image $\boldsymbol{x}$ of size $H \times W \times 3$.

The second loss term is based on an auxiliary adversarial convolutional network. This loss term is large if the adversarial network can discriminate the output of the segmentation network from ground-truth label maps. Since the adversarial CNN has a field-of-view that is either the entire image or a large portion of it, mismatches in the higher-order label statistics can be penalized by the adversarial loss term. Higher-order label statistics (such as *e.g.* the shape of a region of pixels labeled with a certain class, or whether the fraction of pixels in a region of a certain class exceeds a threshold) are not accessible by the standard per-pixel factorized loss function. We use $a(\boldsymbol{x}, \boldsymbol{y}) \in [0, 1]$ to denote the scalar probability with which the adversarial model predicts that $\boldsymbol{y}$ is the ground truth label map of $\boldsymbol{x}$, as opposed to being a label map produced by the segmentation model $s(\cdot)$.

Given a data set of $N$ training images $\boldsymbol{x}_n$ and a corresponding label maps $\boldsymbol{y}_n$, we define the loss as

$$\ell(\boldsymbol{\theta}_s, \boldsymbol{\theta}_a) = \sum_{n=1}^{N} \ell_{\text{mce}}(s(\boldsymbol{x}_n), \boldsymbol{y}_n) - \lambda \Big[ \ell_{\text{bce}}(a(\boldsymbol{x}_n, \boldsymbol{y}_n), 1) + \ell_{\text{bce}}(a(\boldsymbol{x}_n, s(\boldsymbol{x}_n)), 0) \Big], \quad (1)$$

where $\boldsymbol{\theta}_s$ and $\boldsymbol{\theta}_a$ denote the parameters of the segmentation model and adversarial model respectively. In the above, $\ell_{\text{mce}}(\hat{\boldsymbol{y}}, \boldsymbol{y}) = - \sum_{i=1}^{H \times W} \sum_{c=1}^{C} y_{ic} \ln \hat{y}_{ic}$ denotes the multi-class cross-entropy loss for predictions $\hat{\boldsymbol{y}}$, which equals the negative log-likelihood of the target segmentation map $\boldsymbol{y}$ represented using a 1-hot encoding. Similarly, we use $\ell_{\text{bce}}(\hat{z}, z) = -\big[z \ln \hat{z} + (1-z) \ln(1-\hat{z})\big]$, the binary cross-entropy loss. We *minimize* the loss with respect to the parameters $\boldsymbol{\theta}_s$ of the segmentation model, while *maximizing* it w.r.t. the parameters $\boldsymbol{\theta}_a$ of the adversarial model.

**Training the adversarial model.** Since only the second term depends on the adversarial model, training the adversarial model is equivalent to *minimizing* the following binary classification loss

$$\sum_{n=1}^{N} \ell_{\text{bce}}(a(\boldsymbol{x}_n, \boldsymbol{y}_n), 1) + \ell_{\text{bce}}(a(\boldsymbol{x}_n, s(\boldsymbol{x}_n)), 0). \quad (2)$$

In our experiments we let $a(\cdot)$ take the form of a CNN. Below, in Section 3.2, we describe several variants for the adversarial network's architecture, exploring different possibilities for the combination of the inputs and the field-of-view of the adversarial network.

**Training the segmentation model.** Given the adversarial network, the training of the segmentation model minimizes the multi-class cross-entropy loss, while at the same time degrading the performance of the adversarial model. This encourages the segmentation model to produce segmentation maps that are hard to distinguish from ground-truth ones for the adversarial model. The terms of the objective function Eq. (1) relevant to the segmentation model are

$$\sum_{n=1}^{N} \ell_{\text{mce}}(s(\boldsymbol{x}_n), \boldsymbol{y}_n) - \lambda \ell_{\text{bce}}(a(\boldsymbol{x}_n, s(\boldsymbol{x}_n)), 0) \quad (3)$$

We follow Goodfellow *et al.* [9], and replace the term $-\lambda \ell_{\text{bce}}(a(\boldsymbol{x}_n, s(\boldsymbol{x}_n)), 0)$ with $+\lambda \ell_{\text{bce}}(a(\boldsymbol{x}_n, s(\boldsymbol{x}_n)), 1)$ when updating the segmentation model in practice. In other words: instead of minimizing the probability that the adversarial predicts $s(\boldsymbol{x}_n)$ to be synthetic label map for $\boldsymbol{x}_n$, we maximize the probability that the adversarial predicts it to be a ground truth map for $\boldsymbol{x}_n$. It is easy to show that $\ell_{\text{bce}}(a(\boldsymbol{x}_n, s(\boldsymbol{x}_n)), 0)$ and $-\ell_{\text{bce}}(a(\boldsymbol{x}_n, s(\boldsymbol{x}_n)), 1)$ share the same set of critical points. The rationale for this modified update is that it leads to a stronger gradient signal when the adversarial makes accurate predictions on the synthetic/ground-truth nature of the label maps. Preliminary experiments confirmed that this is indeed important in practice to speedup training.

### 3.2 Network architectures

We now detail the architectures we used for our preliminary experiments on the Stanford Background dataset and large-scale experiments on the PASCAL VOC 2012 segmentation benchmark.

**Stanford Background dataset.** For this dataset we used the multi-scale segmentation network of Farabet *et al*. [7], and train it patch-wise from scratch. The adversarial takes as input a label map, and the corresponding RGB image. The label map is either the ground truth corresponding to the image, or produced by the segmentation net. The ground truth label maps are down-sampled to match the output resolution of the segmentation net, and fed in a 1-hot encoding to the adversarial. The architecture of the adversarial is similar to that illustrated in Figure 1, its precise details are given in the supplementary material. At first, two separate branches process the image and the label map, to allow different low level representations for the two different signals. We follow the observation of Pinheiro *et al*. [24] that it is preferable to have roughly the same number of channels for each input signal, so as to avoid that one signal dominates the other when fed to subsequent layers. When fusing the two signal branches, we represent both inputs using 64 channels. The signals are then passed into another stack of convolutional and max-pooling layers, after which the binary class probability is produced by a sigmoid activation. The adversarial network applies two max-pooling operators to the label maps, resulting in a number synthetic/ground-truth predictions of the adversarial that is $4 \times 4 = 16$ times smaller than the number of predictions generated by the segmentation network.

**Pascal VOC 2012 dataset.** For this dataset we used the state-of-the-art Dilated-8 architecture of Yu *et al*. [34], and fine-tune the pre-trained model. This architecture is built upon the VGG-16 architecture [31], but does not include the two last max-pooling layers to maintain a higher resolution. The convolutions that follow the modified pooling operators are dilated with a factor of two for each preceding suppressed max-pooling layer. Following the last convolutional layer, a "context module" composed of eight convolutional layers with increasing dilation factors, is used to expand the network's field-of-view while maintaining the resolution of the feature maps. We explore three variants for the adversarial network input, which we call respectively *Basic*, *Product* and *Scaling*.

In the first approach, *Basic*, we directly input the probability maps generated by the segmentation network. Preliminary experiments in this set-up show no difference when adding the corresponding RGB image, we therefore do not use it for simplicity. One concern for this choice of the inputs is that the adversarial network can potentially trivially distinguish the ground truth and generated label maps by detecting if the map consists of zeros and ones (one-hot coding of ground truth), or of values between zero and one (output of segmentation network).

In the second case, *Product*, we use the label maps to segment the input RGB image, and use it as input for the adversarial. In particular, we multiply the input image with each of the class probability maps (or ground truth), leading to an adversarial input with $3C$ channels. See Figure 2 for illustration.

In the third case, *Scaling*, we replace the 1-hot coding of the ground-truth label maps $y$ with distributions over the labels $\overline{y}$ that put at least mass $\tau$ at the correct label, but are otherwise as similar as possible (in terms of KL divergence) to the distributions produced by the segmenting network. For each spatial position $i$, given its ground-truth label $l$, we set the probability for that pixel and that label to be $\overline{y}_{il} = \max(\tau, s(\boldsymbol{x})_{il})$, where $s(\boldsymbol{x})_{il}$ is the corresponding prediction of the segmentation net. For all other labels $c$ we set $\overline{y}_{ic} = s(\boldsymbol{x})_{ic}(1 - \overline{y}_{il})/(1 - s(\boldsymbol{x})_{il})$, so that the label probabilities in $\overline{y}$ sum to one for each pixel. In our experiments we have used $\tau = 0.9$.

We also need to handle the presence of unlabeled pixels in the ground-truth for the input to the adversarial. We adopt an approach similar to what is done in image-wise training of the segmentation model with the cross entropy loss. We zero-out the values at the spatial positions of unlabeled pixels in both the ground-truth and the output of the segmentation network. We also zero-out the corresponding gradient values during the backward pass corresponding to the second term of Eq. 3. Indeed, those gradients do not correspond to predictions produced by the segmentation net, but to the presence of zeros introduced by this procedure, and should therefore be ignored.

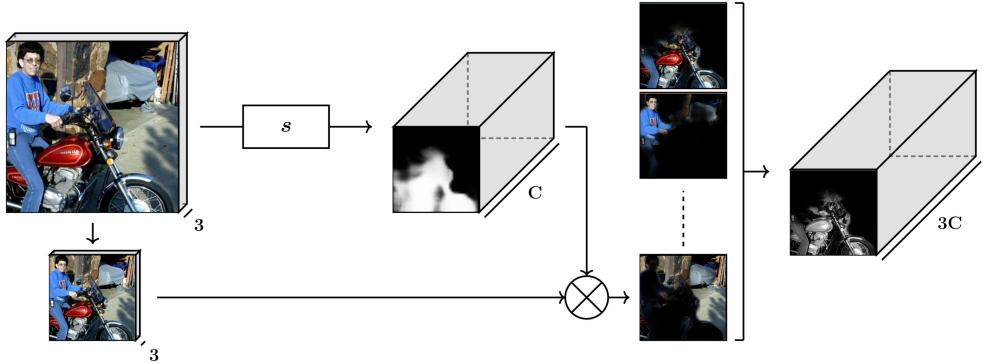

Figure 2: Illustration of using the product of the RGB input and the output of the segmentation network to generate input for the adversarial network. The image is down-sampled by the stride of the segmentation network. The probability maps are then multiplied element-wise with each color channel. These outputs are concatenated and form the input to the adversarial network.

We experiment with two architectures for the adversarial with different fields-of-view. The first architecture, we call *LargeFOV* has a field-of-view of $34 \times 34$ pixels in the label map, whereas the second one, *SmallFOV*, has a field-of-view of $18 \times 18$. Note that this corresponds to a larger image region since the outputs of the segmentation net are eight times down-sampled with respect to the input image. We expect LargeFOV to be more effective to detect differences in patterns of relative position and co-occurrence of class labels over lager areas. Whereas we expect SmallFOV to focus on more fine local details, such as the sharpness and shape of class boundaries and spurious class labels.

Finally, we test a high capacity variant as well as a lighter one of each architecture, the latter one having less channels per layer. All architectures are detailed in the supplementary material.

## 4 Experimental evaluation results

**Datasets.** In our experiments we used two datasets. The Stanford Background dataset [10] contains 715 images of eight classes of scene elements. We used the splits introduced in [10]: 573 images for training, 142 for testing. We train the multi-scale network of Farabet *et al*. [7] using the same hyper-parameters as in [7]. We have further split the training set into eight subsets, and we train on all subsets but one, which we use as our validation set to choose an appropriate weight $\lambda$, learning rate for the adversarial network and to select the final model. The adversarial network is trained using a weight $\lambda = 2$ and learning rate $10^{-3}$. We compute the three standard performance measures: per class accuracy, per pixel accuracy, and the mean Intersection over Union (IoU) as defined in [6].

The second dataset is Pascal VOC 2012. As is common practice, we train our models on the dataset augmented with extra annotations from [12], which gives a total of $10,582$ training images. For validation and test, we use the challenge's original $1,449$ validation images and $1,456$ test images.

In addition to the standard IoU metric, we also evaluate our models using the BF measure introduced by [3], to measure accuracy along object contours. This measure extends the Berkeley contour matching score [17], a commonly used metric in segmentation, to semantic segmentation. It is based on the closest match between boundary points in the prediction and the ground-truth segmentation. The tolerance in the distance error, used to decide whether a point has a match or not, is a factor $\theta$ times the length of the image diagonal. We choose $\theta$ such that this distance error tolerance is 5 pixels for the smallest image diagonal. In the original annotations of the dataset, however, the labels around the border of the objects are not given, since they are marked as 'void' and ignored in evaluation. Instead, to measure the mean BF, we use the 1,103 images out of 1,449 images of the validation set which were annotated on all pixels by [12].

**Results on the Stanford Background dataset.** In Figure 3 we give an illustration of the segmentations generated using this network with and without adversarial training. The adversarial training better enforces spatial consistency among the class labels. It smoothens and strengthens the class probabilities over large areas, see *e.g.* the probability maps for *sky* and *grass*, but also sharpens class boundaries, and removes spurious class labels across small areas.

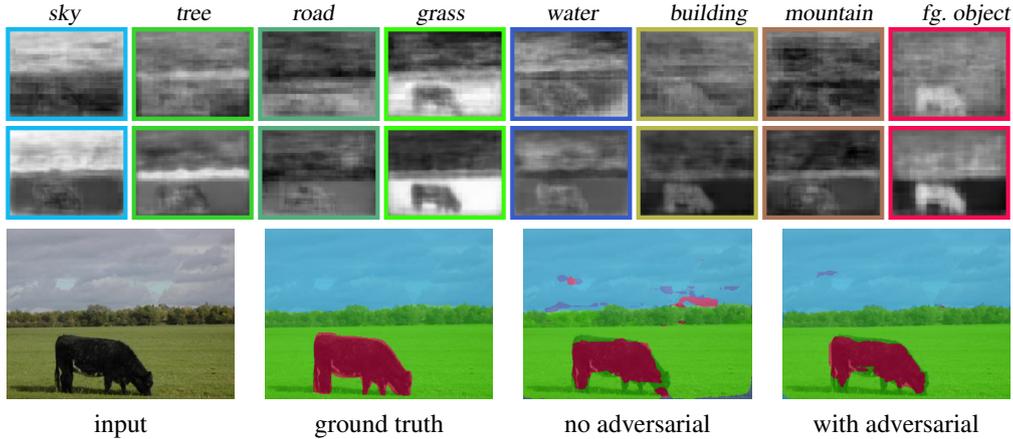

Figure 3: Segmentations on Stanford Background. Class probabilities without (first row) and with (second row) adversarial training. In the last row the class labels are superimposed on the image.

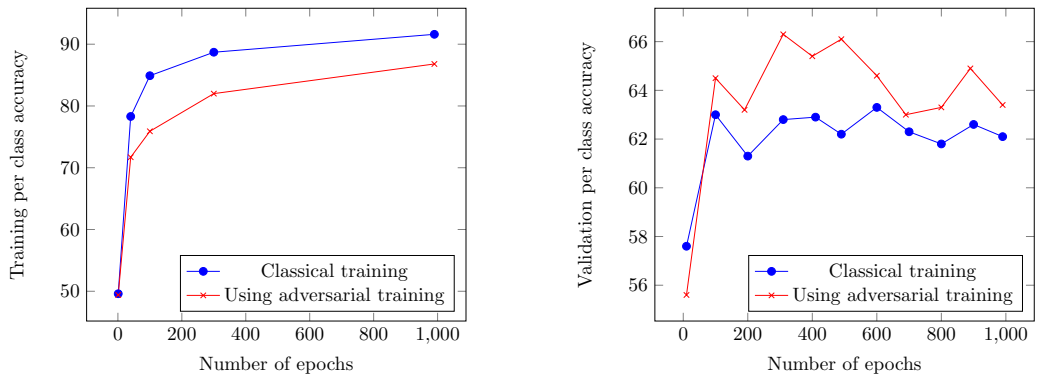

Figure 4: Per-class accuracy across training ephochs on the Stanford Background dataset on train data (left) and validation data (right), with and without adversarial training.

In Figure 4 we display the evolution of the per-class prediction accuracy on the train and validation sets, using either standard or adversarial training. Note that the adversarial strategy results in less overfitting, *i.e.* generating a regularization effect, resulting in improved accuracy on validation data. This is also reflected in all three performance metrics on the test set, as reported in 1.

**Results on PASCAL VOC 2012.** In order to set the learning rates of both the segmentation and adversarial network, as well as the trade-off weight of the losses $\lambda$, we conduct a small grid search for each combination of adversarial architecture and input encoding.

To train the two networks, we first experimented with pre-training the adversarial network before using the adversarial loss to fine-tune the segmentation network, so as to ensure that the adversarial loss is meaningful. This, however, led to the training to be rapidly instable after just a few epochs in many experiments. We found that training instead using an alternating scheme is more effective. We experimented with a *fast* alternating scheme, where we alternate between updating the segmenting network's and the adversarial network's weights at every iteration of SGD and a *slow* one, where we alternate only after 500 iterations of each. We found the second scheme to led to the most stable training, and used it for the results reported in Table 2. For details on the hyper-parameter search, and the final hyper-parameters used for each model, we refer the reader to the supplementary material.

|  | Per-class acc. | Pixel acc. | Mean IoU |
|---|---|---|---|
| Standard | 66.5 | 73.3 | 51.3 |
| Adversarial | **68.7** | **75.2** | **54.3** |

Table 1: Segmentation accuracy on the Stanford Background dataset.

|  | Basic | | Product | | Scaling | |
| --- | --- | --- | --- | --- | --- | --- |
|  | mIOU | mBF | mIOU | mBF | mIOU | mBF |
| LargeFOV | **72.0** | 47.2 | **72.0** | 47.7 | **72.0** | 47.9 |
| SmallFOV | **72.0** | **47.6** | 71.9 | 46.4 | 71.9 | 47.1 |
| LargeFOV-light | **72.0** | 47.0 | **72.0** | 47.7 | **72.0** | 47.4 |
| SmallFOV-light | 71.9 | 47.2 | 71.9 | 47.4 | **72.0** | 46.9 |

Table 2: Performance using different architectures and input encodings for the adversarial model.

We compare the results of adversarial training with a baseline consisting of fine-tuning of Dilated8 using the cross-entropy loss only. For the baseline we obtained a mean IoU of 71.8 and mean BF of 47.4. As shown in Table 2, we observe small but consistent gains for most adversarial training setups. In particular, the LargeFOV architecture is the most effective overall. Moreover, it is interesting to note that the different adversarial input encodings lead to comparable results. In fact, we found that for the basic input encoding, the adversarial does not succeed in perfectly separating ground-truth and predicted label maps, it rather has a discrimination accuracy that is comparable to that obtained with the other input encodings.

Using the evaluation server we also tested selected models on the PASCAL VOC 2012 test set. For the baseline model we obtain (73.1), while for LargeFOV-Product and LargeFOV-Scaling we obtained 73.3 and 73.2 respectively. This confirms the small but consistent gains that we observed on the validation data.

## 5 Discussion

We have presented an adversarial approach to learn semantic segmentation models. In the original work of Goodfellow *et al*. [9] the adversarial model is used to define a proxy loss for a generative model in which the calculation of the cross-entropy loss is intractable. In contrast, the CNN-based segmentation models we use allow for tractable computation of the exact multi-class cross-entropy loss. In our work we use the adversarial network as a "variational" loss, with adjustable parameters, to regularize the segmentation model by enforcing higher-order consistency in the factorized prediction model of the label variables. Methodologically, this approach is related to work by Roweis *et al*. [28], where variational inference was used in tractable linear-Gaussian mixture models to enforce consistency across multiple local dimension reduction models, and to work by Tarlow and Zemel [33] which learn models with higher-order loss terms, while not including such higher-order terms in the predictive model to ensure efficient inference.

To demonstrate the regularization property of adversarial training, we conducted experiments on the Standford Background dataset and the PASCAL VOC 2012 dataset. Our results show that the adversarial training approach leads to improvements in semantic segmentation accuracy on both datasets. The gains in accuracy observed on the Stanford Background dataset are more pronounced. This is most likely due to higher risk of over fitting using this smaller data set, and also due to the more powerful segmentation architectures used for the PASCAL VOC 2012 dataset.

**Acknowledgments**

This work has been partially supported by the LabEx PERSYVAL-Lab (ANR-11-LABX-0025-01).

# Semantic Segmentation using Adversarial Networks
## — Supplementary Material —

**Pauline Luc**
Facebook AI Research
Paris, France
paulineluc@fb.com

**Camille Couprie**
Facebook AI Research
Paris, France
coupriec@fb.com

**Soumith Chintala**
Facebook AI Research
New York, USA
soumith@fb.com

**Jakob Verbeek**
INRIA, Laboratoire Jean Kuntzmann
Grenoble, France
firstname.lastname@inria.fr

# 1 Network architectures

**Stanford background dataset.** The adversarial network architecture we used in this case is detailed in Figure 1. We apply local contrast normalization to the RGB images before entering them into either the adversarial or segmentation network.

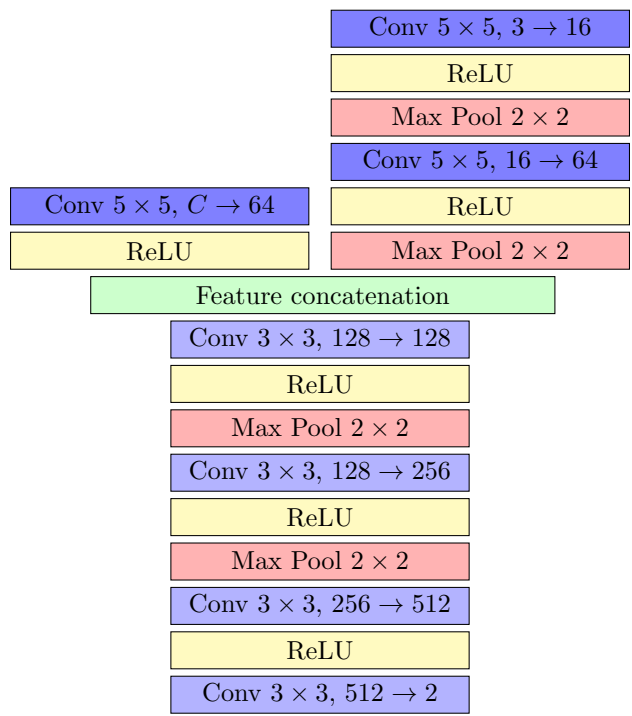

Figure 1: Adversarial architecture for the experiments with the Stanford Background dataset. The left and right branch processes respectively the class segmentations and the RGB image.

**PASCAL VOC 2012 dataset.** For PASCAL VOC 2012, we have experimented with two architectures for the adversarial network with different fields-of-view, LargeFOV and SmallFOV. We have also tested for each architecture a smaller capacity variant, respectively called LargeFOV-light and SmallFOV-light. Table 1 summarizes the architectures used. The number of parameters of each model also depends on the input encoding. In the *Product* adversarial input scheme, the RGB image is mean-centered per pixel, in the same way as before being input to the segmenting model.

| LargeFOV-light | LargeFOV | SmallFOV-light | SmallFOV |
|---|---|---|---|
| Conv $3\times 3$ | $C' \to 96$ | $C' \to 96$ | |
| ReLU | | | |
| Conv $3\times 3$ | $96 \to 128$ | $96 \to 128$ | |
| ReLU | | | |
| Conv $3\times 3$ | $128 \to 128$ | $128 \to 128$ | |
| ReLU | | | |
| Max Pool $2\times 2$ | | | |
| Conv $3\times 3$ | $128 \to 128$ | $128 \to 256$ | |
| ReLU | | | |
| Conv $3\times 3$ | $128 \to 128$ | $256 \to 256$ | |
| ReLU | | | |
| Max Pool $2\times 2$ | | | |
| Conv $3\times 3$ | $128 \to 256$ | $256 \to 512$ | |
| ReLU | | | |
| Conv $3\times 3$ | $256 \to 2$ | $512 \to 2$ | |
| $C' = C$ | $8,7.10^5$ | $2,4.10^6$ | |
| $C' = 3C$ | $9,1.10^5$ | $2,4.10^6$ | |

| SmallFOV-light | | SmallFOV |
|---|---|---|
| Conv $3\times 3$ | $C' \to 96$ | $C' \to 96$ |
| ReLU | | |
| Conv $1\times 1$ | $96 \to 128$ | $96 \to 128$ |
| ReLU | | |
| Max Pool $2\times 2$ | | |
| Conv $3\times 3$ | $128 \to 128$ | $128 \to 256$ |
| ReLU | | |
| Conv $1\times 1$ | $128 \to 128$ | $256 \to 256$ |
| ReLU | | |
| Max Pool $2\times 2$ | | |
| Conv $3\times 3$ | $128 \to 256$ | $256 \to 512$ |
| ReLU | | |
| Conv $1\times 1$ | $256 \to 2$ | $512 \to 2$ |
| $C' = C$ | $4,9.10^5$ | $2,4.10^6$ |
| $C' = 3C$ | $9,1.10^5$ | $2,4.10^6$ |

Table 1: Summary of the architectures used for the adversarial network, from left to right : LargeFOV-light, LargeFOV, SmallFOV-light, SmallFOV, with layers organized from top to bottom, along with the approximate number of parameters for each model. This number depends on the number of channels of the input ($C$ channels for *Basic* and *Scaling* encodings, $3C$ channels for *Product* encoding.)

## 2 Additional results

For our baseline which does not use adversarial training, we fine-tuned the Dilated8 segmenting architecture with learning rates $10^{-5}, 10^{-6}, 10^{-7}$. We obtained best results using $10^{-6}$.

For the adversarial net we tested learning rates ($alr$) in the range $[0.01, 0.5]$. For the segmentation net we tested learning rates ($slr$) in the range $[10^{-6}, 10^{-4}]$. For $\lambda$, we tested values 0.1 and 1. We report results across all adversarial architectures and input encodings in Table 2, together with the selected hyper-parameters for each setting. For the BF measure we also report the standard deviation $\sigma$ across the images.

We report the per-class accuracy on the test set for selected models in Table 3.

|  | mIOU |  |  | mBF |  |  | $\sigma$ |  |  | $slr$ |
|---|---|---|---|---|---|---|---|---|---|---|
| Baseline | 71.79 |  |  | 47.36 |  |  | 22.21 |  |  | $10^{-6}$ |

| Architecture | Simple | | | Product | | | Scaling | | |
|---|---|---|---|---|---|---|---|---|---|
|  | mIOU | mBF | $\sigma$ | mIOU | mBF | $\sigma$ | mIOU | mBF | $\sigma$ |
|  | $slr$ | $alr$ | $\lambda$ | $slr$ | $alr$ | $\lambda$ | $slr$ | $alr$ | $\lambda$ |
| LargeFOV | **72.02** | 47.19 | 21.67 | **72.04** | 47.70 | 22.42 | **71.99** | 47.85 | 22.25 |
|  | $10^{-4}$ | 0.5 | 1 | $10^{-6}$ | 0.1 | 1 | $10^{-5}$ | 0.1 | 1 |
| SmallFOV | 71.95 | 47.58 | 22.20 | 71.87 | 46.42 | 21.88 | 71.89 | 47.14 | 21.80 |
|  | $10^{-5}$ | 0.2 | 1 | $10^{-4}$ | 0.1 | 0.1 | $10^{-4}$ | 0.5 | 1 |
| LargeFOV-light | 71.95 | 46.97 | 21.96 | **71.97** | 47.73 | 22.32 | **72.01** | 47.43 | 22.17 |
|  | $10^{-4}$ | 0.5 | 0.1 | $10^{-6}$ | 0.1 | 1 | $10^{-4}$ | 0.1 | 0.1 |
| SmallFOV-light | 71.88 | 47.22 | 21.99 | 71.91 | 47.38 | 22.39 | **71.96** | 46.89 | 21.71 |
|  | $10^{-5}$ | 0.1 | 1 | $10^{-5}$ | 0.2 | 1 | $10^{-4}$ | 0.5 | 1 |

Table 2: Performance and hyper-parameters of the baseline model (no adversarial training, top), and various adversarial architectures and input encodings.

| | background | aeroplane | bicycle | bird | boat | bottle | bus | car | cat | chair | cow | dining table | dog | horse | motorbike | person | potted plant | sheep | sofa | train | tv monitor | mIOU |
|---|---|---|---|---|---|---|---|---|---|---|---|---|---|---|---|---|---|---|---|---|---|---|
| Baseline | 92.9 | **87.2** | 38.2 | 84.5 | 62.3 | 69.7 | 88.0 | 82.3 | 86.4 | 34.6 | **80.5** | 60.5 | **81.0** | **86.3** | 83.0 | **82.7** | 53.6 | **83.9** | 54.5 | 79.3 | **63.7** | 73.1 |
| LargeFOV-Product | 93.0 | 87.1 | **38.5** | **84.9** | 63.2 | 69.7 | 88.0 | **82.5** | 86.8 | 34.5 | 80.3 | **61.5** | 80.9 | 85.8 | **83.3** | 82.6 | **55.0** | 83.5 | **54.7** | 79.5 | 62.9 | **73.3** |
| LargeFOV-Scaling | **94.9** | 87.1 | 38.3 | 84.8 | **63.3** | **69.8** | **88.1** | 82.2 | **87.0** | **34.7** | 80.4 | 60.4 | **81.0** | 86.1 | 83.1 | **82.7** | 54.5 | 83.4 | 53.9 | 79.5 | 63.4 | 73.2 |

Table 3: Results on the Pascal VOC test set.

3